\documentclass[letterpaper]{article} 
\usepackage{aaai2026}  
\usepackage{times}  
\usepackage{helvet}  
\usepackage{courier}  
\usepackage[hyphens]{url}  
\usepackage{graphicx} 
\usepackage{natbib}  
\usepackage{caption} 
\usepackage{color}
\definecolor{softgreen}{RGB}{92, 184, 92}
\usepackage{booktabs}
\usepackage{multirow} 
\usepackage{amsmath}
\usepackage{breqn} 
\usepackage{cleveref}
   
\usepackage{makecell}

\usepackage[ruled,vlined]{algorithm2e}
\usepackage{amssymb}

\usepackage{newfloat}
\usepackage{listings}
\DeclareCaptionStyle{ruled}{labelfont=normalfont,labelsep=colon,strut=off} 
\usepackage{amsmath, amssymb}
\usepackage{booktabs}
\usepackage{listings}
\usepackage{xcolor}
\definecolor{softgreen}{RGB}{102,205,170}
\usepackage{multirow} 
\title{Privacy-protected Retrieval-Augmented Generation for \\Knowledge Graph Question Answering}
\author {
    Yunfeng Ning\textsuperscript{\rm 1, \thanks{Yunfeng Ning and Mayi Xu contribute equally to this work.}},
    Mayi Xu\textsuperscript{\rm 1, $^{*}$}, 
    Jintao Wen\textsuperscript{\rm 1}, 
    Qiankun Pi\textsuperscript{\rm 1},
    Yuanyuan Zhu\textsuperscript{\rm 1}, 
    Ming Zhong\textsuperscript{\rm 1},\\
    Jiawei Jiang\textsuperscript{\rm 1, \thanks{Corresponding authors: Tieyun Qian and Jiawei Jiang.}},
    Tieyun Qian\textsuperscript{\rm 1, $^\dagger$}
}
\affiliations {
    \textsuperscript{\rm 1}School of Computer Science, Wuhan University, China\\
    \{ningyunfeng, xumayi,  jiawei.jiang, qty\}@whu.edu.com
}

\usepackage{bibentry}

\begin{document}

\maketitle

\begin{abstract}

Large Language Models (LLMs) often suffer from hallucinations and outdated or incomplete knowledge. Retrieval-Augmented Generation (RAG) is proposed to address these issues by integrating external knowledge like that in knowledge graphs (KGs) into LLMs. However, leveraging private KGs in RAG systems poses significant privacy risks due to the black-box nature of LLMs and potential insecure data transmission.
In this paper, we investigate the \emph{privacy-protected RAG scenario} for the first time, where entities in KGs are anonymous for LLMs, thus preventing them from accessing entity semantics. Due to the loss of semantics of entities, previous RAG systems cannot retrieve question-relevant knowledge from KGs by matching questions with the meaningless identifiers of anonymous entities. To realize an effective RAG system in this scenario, two key challenges must be addressed: (1) \emph{How can anonymous entities be converted into retrievable information}? (2) \emph{How to retrieve question-relevant anonymous entities}?

To address these challenges, we propose a novel \textbf{A}bstraction \textbf{R}easoning \textbf{o}n \textbf{G}raph (\textbf{ARoG}) framework including relation-centric abstraction and structure-oriented abstraction strategies. For challenge (1), the first strategy abstracts entities into high-level concepts by dynamically capturing the semantics of their adjacent relations. Hence, it supplements meaningful semantics which can further support the retrieval process. For challenge (2), the second strategy transforms unstructured natural language questions into structured abstract concept paths. These paths can be more effectively aligned with the abstracted concepts in KGs, thereby improving retrieval performance. In addition to guiding LLMs to effectively retrieve knowledge from KGs, these abstraction strategies also strictly protect privacy from being exposed to LLMs. Experiments on three datasets demonstrate that ARoG achieves strong performance and privacy-robustness, establishing a new practical direction for privacy-protected RAG systems. 
\begin{links}
    \link{Code}{https://github.com/NLPGM/ARoG}
\end{links}

\end{abstract}

\section{Introduction}
\begin{figure}[t]
  \centering
  \includegraphics[width=\columnwidth]{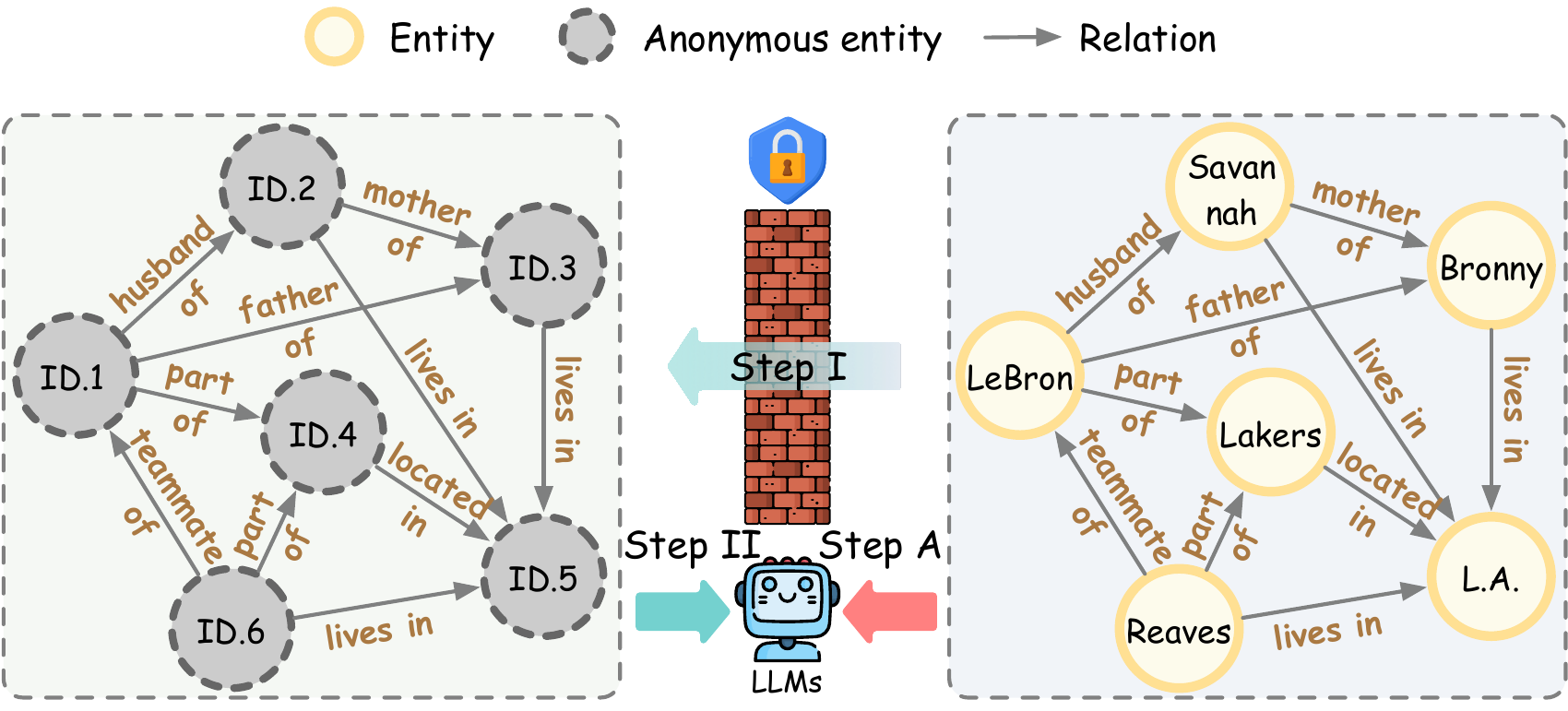}
  \caption{Comparison between previous RAG (Step A) with privacy-protected RAG (Step I-II) system for KGQA.}

  \label{fig:fig1}
\end{figure}

Large Language Models (LLMs) exhibit remarkable capabilities across a wide range of natural language tasks \citep{Wang_Hu_He_Xu_Liu_Liu_Shen_2024}. However, their practical deployment remains constrained by critical limitations, including factual inaccuracies (hallucinations) and outdated or incomplete knowledge \citep{ji_hallucination_survey}. 
To address these issues, the Retrieval-Augmented Generation (RAG) technique has been proposed to first retrieve factual knowledge from external sources and then incorporate them into LLMs during inference, thereby enhancing the factual grounding and reliability of LLM outputs \citep{shi-etal-2024-replug}.

KGs can store a large amount of factual knowledge. 
Hence, they are widely used in the RAG system as external knowledge sources to provide supplementary knowledge to LLMs \citep{sun2024thinkongraph}. The effectiveness of KG-enhanced RAG systems is commonly evaluated through Knowledge Graph Question Answering (KGQA) task. Specifically, given a question, the RAG system aims first to retrieve relevant factual triplets from the KGs and then generate answers based on them.

In practice, many private KGs contain
sensitive information like personal data and confidential company details \citep{10272537}. 
When using such private KGs to answer questions requiring privacy details, current RAG systems must retrieve relevant privacy data and expose them to the LLMs. As shown in the Step A of Figure~\ref{fig:fig1}, when retrieving triplets to answer the question ``Where does Bronny live?'', the factual triplet $(Bronny, lives\ in, L.A.)$ containing
sensitive information will be exposed to LLMs. 
The inherent black-box nature of LLMs and potential insecure data transmission between users and LLM servers highlight the urgent need for robust privacy protections. This need is even greater when using third-party LLM APIs, as users may lack transparency or direct control over how their 
private data are collected, stored, or processed.

In this paper, we make the first attempt to address privacy concerns when the RAG system retrieves external knowledge from KGs. In this \emph{privacy-protected RAG scenario}, as depicted in the Step I of Figure~\ref{fig:fig1}, entities within the KG are anonymous for LLMs and replaced with corresponding unique Machine Identifiers (MIDs), which are encrypted and contain no semantic content. 
As a result, as illustrated in the Step II of Figure~\ref{fig:fig1}, the LLMs are unable to access the semantic information (e.g., the specific types, names, and descriptions) of these entities.

Due to the loss of semantics of entities, previous RAG systems cannot retrieve question-relevant knowledge from KGs by matching questions with the meaningless MIDs. To realize an effective RAG system in this scenario, two key challenges must be addressed: (1) \emph{How can anonymous entities be converted into retrievable information}? (2) \emph{How to retrieve question-relevant anonymous entities}?

To address these challenges, we propose a novel \textbf{A}bstraction \textbf{R}easoning \textbf{o}n \textbf{G}raph (\textbf{ARoG}) framework including relation-centric abstraction and structure-oriented abstraction strategies. 

For challenge (1), the relation-centric abstraction strategy treats the anonymous entities as subject or object nouns and their adjacent relations as predicate verbs, then abstracts the entities into corresponding higher-level concepts. These concepts are generated by LLMs based on the adjacent relations of entities. This strategy captures their in-context semantics and then appends to the MIDs, thereby overcoming challenge (1). For instance, an entity that serves as the subject of relations ``time\_zones'', ``contained\_by'', ``population'' and the object of relation ``citytown'' can be abstracted into ``geographic location''.

For challenge (2), the structure-oriented abstraction strategy transforms unstructured natural language questions into structured abstract concept paths. These paths can be more effectively aligned with the abstracted concepts in KGs than naive questions, thereby improving retrieval performance. For instance, the abstract concept path to the question ``What is the name of the daughter of the artist who had the The Mrs. Carter Show World Tour?'' is ``Nicki Minaj (artist) $\rightarrow$ had $\rightarrow$ The Mrs. Carter Show World Tour; Nicki Minaj (artist) $\rightarrow$ has daughter named $\rightarrow$ Chiara Fattorini (person)''. 
Notably, the semantic similarity between this path and relevant triplets persists even with inaccurate entities, as the path captures the relational structure of abstracted concepts.

To fulfill the complete privacy-protected RAG, our ARoG framework also incorporates an Abstraction-driven Retrieval module and a Generator module. These two modules ensure robust performance in retrieving question-relevant triplets from KGs and generating answers.
To evaluate the effectiveness of our ARoG framework, we conduct experiments in the \emph{privacy-protected RAG scenario} on three popular yet diverse datasets including WebQSP, CWQ, and GrailQA. Our ARoG framework achieves the state-of-the-art (SoTA) performance on all three datasets. 

Our contributions can be summarized as follows.
\begin{itemize}
\item We explore the \emph{privacy-protected RAG scenario} for the first time, which aims to address privacy concerns when the RAG system retrieves external knowledge from KGs.
\item We propose a novel RAG-based framework ARoG, utilizing two abstraction strategies to tackle challenges in the \emph{privacy-protected RAG scenario} while balancing high performance and data privacy.
\item We conduct extensive experiments across three public datasets, providing empirical evidence of the effectiveness and robustness of our proposed ARoG.
\end{itemize}

\section{Related Work}
In this section, we review related works for the KGQA task.
\subsection{Semantic Parsing for KGQA}
Semantic parsing (SP)-based methods treat LLMs as semantic parsers that translate questions into formal queries using labeled exemplars. To address errors in the initial formal queries, \citet{li-etal-2023-shot} utilize entity and relation binders grounded in KGs, \citet{Nie_Zhang_Wang_Liu_2024} reformulate the generation process into code generation process; and \citet{Zhang_Jin_Zhu_Chen_Huang_Wang_Hua_Liang_Chen_2025} adapt generation process to function calling using Condition Graphs (CGs), a variant of KGs. However,these methods do not fundamentally resolve the challenges posed by formatting errors, and the performance is heavily limited by the quality and quantity of labeled exemplars.

\subsection{Retrieval-Augmented Generation for KGQA}
Compared with SP-based methods, RAG-based methods usually have superior performance. RAG-based methods employ LLMs as retrievers to fetch question-relevant triplets as evidence, and employ LLMs as generators to infer answers with evidence. To improve the quality of evidence, \citet{jiang-etal-2023-structgpt} propose an iterative LLM-based ``Read-then-Reason'' framework; \citet{sun2024thinkongraph} utilize LLMs to perform beam search on KGs; and \citet{xu-etal-2024-generate} utilize LLMs to supplement additional knowledge. However, previous RAG-based KGQA methods directly feed factual triplets within KGs to LLMs, which may undermine data privacy. The inherent black-box nature of LLMs and potential insecure data transmission between users and LLM servers highlight the urgent need for robust privacy protections. This need is even greater when using third-party LLM APIs, as users may lack transparency or direct control over how their privacy data are collected, stored, or processed.

In this paper, we make the first attempt to explore the \emph{privacy-protected RAG scenario} for KGQA, which aims to address privacy concerns when the RAG system retrieves external knowledge from KGs.
\begin{figure*}[t]
  \centering
 \includegraphics[width=\textwidth]{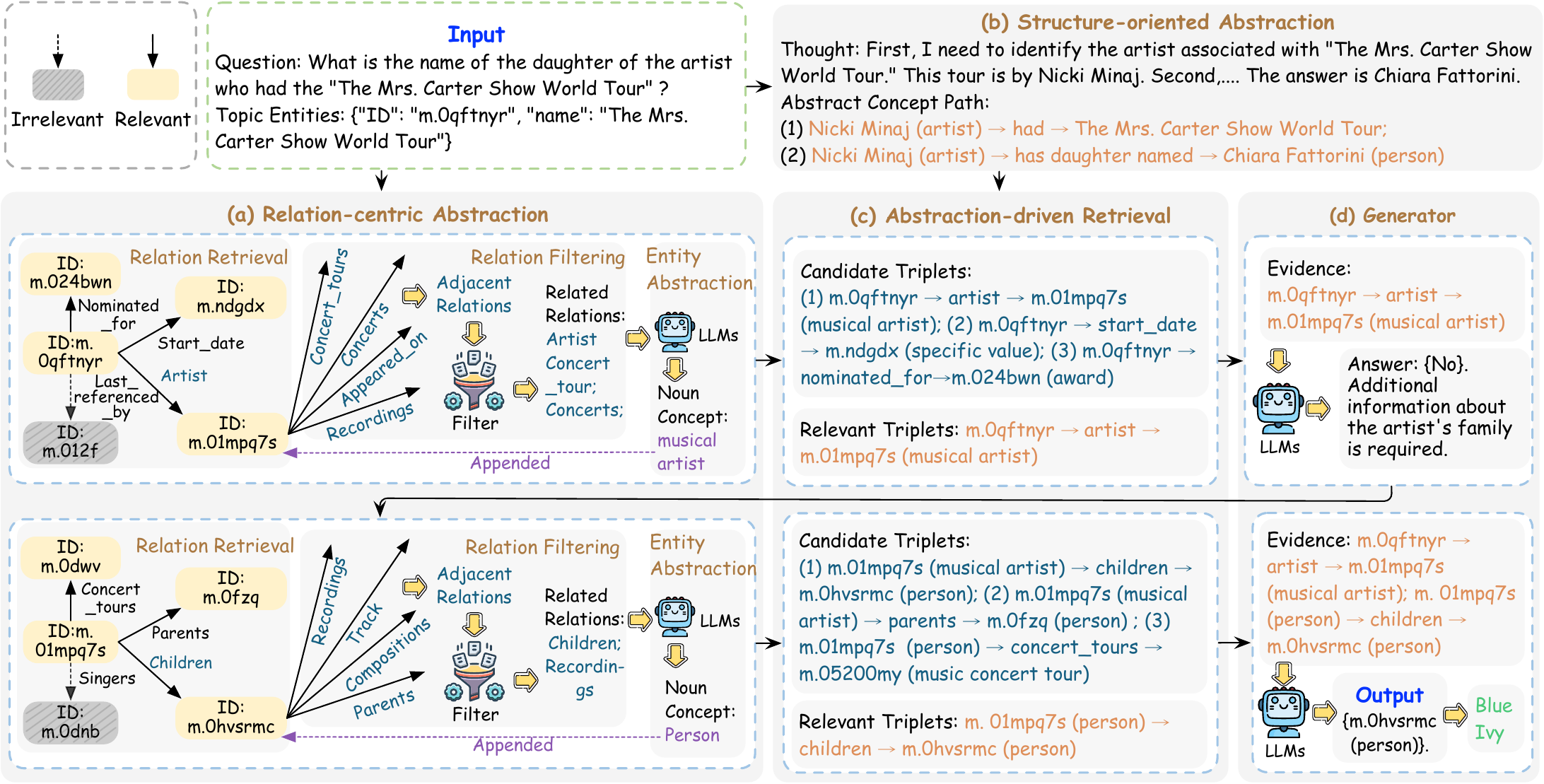}
  \caption{An overview of ARoG framework, which includes four primarily modules: Relation-centric Abstraction, Structure-oriented Abstraction, Abstraction-driven Retrieval and Generator modules. The Relation-centric Abstraction module consists of three steps: relation retrieval, relation filtering and entity abstraction.}
  \label{fig:fig2}
\end{figure*}

\section{Methodology}

\subsection{Preliminary}
\paragraph{Knowledge Graphs (KGs)} Knowledge Graphs (KGs) contain abundant factual triplets (${e_h}, r , e_t$), that is, $KG=\{ (e_h, r , e_t) \mid e_h, e_t \in \mathcal{E}, r \in \mathcal{R} \}$, where $e_h$, $r$, and $e_t$ represent the head entity, the relation, and the tail entity, respectively. $\mathcal{E}$ and $\mathcal{R}$ denote the set of entities and relations, respectively.

\paragraph{Knowledge Graph Question Answering (KGQA)} Given an nature language question $q$, the topic entities $E_q\ =\ \left\{e_q^1,e_q^2,\cdot\cdot\cdot,e_q^n\right\}$ (the main entities asked in the question \citep{10.1145/3184558.3186916}), and a knowledge graph $KG$, the task of KGQA is to predict the answer entities $A_q\ =\ \left\{e_a^1,e_a^2,\cdot\cdot\cdot,e_a^n\right\}$, which can answer the question. Following previous works \citep{sun2024thinkongraph}, we assume that the topic entities and answer entities are labeled and linked to the corresponding entities in $KG$, i.e., $ E_q,A_q\subseteq \mathcal{E}$. 

\paragraph{{Privacy-protected RAG scenario} for KGQA} 
To prevent the LLM from accessing sensitive information within the KG, we anonymize all entities by representing them as MIDs.
Specifically, we leverage the Freebase database system \citep{Bollacker-etal-2008-freebase} to map each entity to a unique, encrypted MID.
These MIDs are meaningless to the LLM. 
Crucially, to maintain task feasibility, the names of the topic entities explicitly mentioned in the question are available.

\subsection{Overview}
In order to solve the challenges in the privacy-protected RAG scenario, we propose a novel Abstraction Reasoning on Graph (ARoG) framework.
The architecture of ARoG, illustrated in Figure \ref{fig:fig2}, follows a Retrieval-then-Generation pipeline. It comprises four primary modules: the first three constitute the retrieval phase, while the latter one belongs to the generation phase, as detailed below.

\textbf{(1) Relation-centric Abstraction (RA):} To overcome the challenge of \emph{How can anonymous entities be converted into retrievable information}, we develop the Relation-centric Abstraction module. It utilizes LLMs to transform anonymous entities into abstract concepts and then appends to the anonymous identifiers to supplement semantic.

\textbf{(2) Structure-oriented Abstraction (SA):} To overcome the challenge of \emph{How to retrieve question-relevant anonymous entities}, we develop the Structure-oriented Abstraction module. It uses LLMs to convert unstructured questions into structured abstract concept paths to guide the retrieval.

\textbf{(3) Abstraction-driven Retrieval (AR):} To retrieve the question-relevant triplets from a large-scale KG, this module sequentially performs searching and pruning operations over the KG, driven by the two aforementioned abstraction modules.

\textbf{(4) Generator:} This module employs an LLM to infer answers based on the question and the question-relevant triplets derived from Abstraction-driven Retrieval.

\subsection{Relation-centric Abstraction Module} \label{Section_4_2}

The Relation-centric Abstraction module treats relations adjacent to entities as verbs and the entities themselves as subject or object nouns. Leveraging the generative capabilities of LLMs, we infer abstract concepts for the entities from these verbs, thereby enriching their semantics. 

As illustrated in Part (a) of Figure \ref{fig:fig2}, the Relation-centric Abstraction module operates on a question $q$, $n$ topic entities $E_q$ and the KG, and comprises three steps as follows.

\paragraph{(1) Relation Retrieval}
First, starting from a set of $n$ topic entities, we explore the KG to extract $n$ sets of adjacent relations. For each relation cluster $R_{t}$, we prompt an LLM to identify the $W$ most question-relevant relations, which are denoted as $R_{t,opt}$. The process is expressed as follows.

{
\small
\begin{equation}
R_{\text{t,opt}} = \mathop{\arg\max}_{\substack{S \subseteq R_{t}, |S| \leq W }} \text{LLM}(R_{t}, q, Inst_{rr}, E_{rr}),
\label{eq:formu4}
\end{equation}
}where $W$ is a hyperparameter representing the width of the retrieval, $Inst_{rr}$ describes the details of the relation retrieval task, $E_{rr}$ denotes an exemplar, and \text{LLM} denotes LLM call.

After retrieval, we gather $n\times W$ entity-relation pairs. 
Subsequently, for each pair, we identify its adjacent entity cluster, resulting in a total of $n\times W$ candidate entity clusters.

\paragraph{(2) Relation Filtering}
Second, while the entities in the candidate clusters are meaningless, their adjacent relations provide rich in-context semantic information. For any entity $e$ in the candidate entity clusters, we treat its adjacent relations as verbs, denoted as $R_{v}$. 
To filter out irrelevant relations, we employ SentenceTransformer \citep{reimers-2020-multilingual-sentence-bert} to select the top-$K$ most relevant relations, denoted as $R_{v,opt}$, based on their cosine similarity to the question $q$.
The process is expressed as follows.
{
\small
\begin{equation}
R_{v,opt} =
\mathop{\arg\max}_{\substack{S \subseteq R_{v}, |S| \leq K}}
\sum_{r \in S} \cos(\text{Emb}(q), \text{Emb}(r)),
\label{eq:formu2}
\end{equation}
}where \text{Emb}(.) represents the embedding results calculated by SentenceTransformer, and $K$ is set to 5 empirically.

\paragraph{(3) Entity Abstraction}
To abstract the representative anonymous entity $e$, we first prompt an LLM to infer its corresponding concept based on the related relations $R_{v,opt}$. We then append this abstract concept to the entity’s MID,  yielding the abstracted entity $e^{abs}$. The procedure is expressed as follows.
{
\small
\begin{equation}
e^{abs}=\text{LLM}(R_{v,opt},e,Inst_{ea},E_{ea}),
\label{eq:formu3}
\end{equation}
}where $Inst_{ea}$ describes the details of the entity abstraction, and $E_{ea}$ denotes a single exemplar for guidance. 

Since the relations within the KG define the schema structures rather than sensitive information, sharing these relations with LLMs poses minimal privacy risk. 
Additionally, to minimize LLM calls, we assume that entities within the same entity cluster share a common concept, in line with \citet{jiang2023unikgqa}. 
If an entity is associated with multiple generated abstract concepts, these concepts are combined into a comma-separated sequence.

This abstraction process is applied to each triple $t=(e_h , r , e_t)$ in the KG, replacing its entities with their abstracted counterparts to form an abstracted triple $t^{abs}=(e_h^{abs} , r , e_t^{abs})$. We then apply this transformation uniformly to the entire set of triplets $T_q=\{t_1,t_2, \cdot\cdot \cdot \}$, yielding its abstracted counterpart, $T_q^{abs}=\{t_1^{abs},t_2^{abs}, \cdot\cdot \cdot\}$.

\subsection{Structure-oriented Abstraction Module} \label{Section_4_3}

The Structure-oriented Abstraction module converts an unstructured question into a structured abstract concept path $P_q$.
During this process, unknown entities are also abstracted into their corresponding concepts. Formally, $P_q$ comprises a set of generated triplets, denoted as $P_q=\{t_{p1},t_{p2},\cdot \cdot \cdot \}$.
By guiding the retrieval of relevant triplets and entities via semantic matching, this path effectively isolates the LLM from the raw entities within the KG.

A key advantage of this strategy is that the effectiveness of abstract concept paths is not reliant on correct entities.
For example, as shown in Part (b) of Figure \ref{fig:fig2}, although the generated entities ``Nicki Minaj'' and ``Chiara Fattorini'' are incorrect, the two resulting triplets, which incorporate the concepts ``artist'' and ``person’', semantically align well with relevant triplets incorporating abstracted entities in the KG.

Specifically, given a question $q$, we utilize an LLM to generate a structured abstract concept path $P_q$. The overall process is expressed as follows.
{
\small
\begin{equation}
Rat,P_q=\text{LLM}(q, Inst_{sa}, E_{sa}),
\label{eq:formu1}
\end{equation}
}where $Inst_{sa}$ describes the abstraction task in a chain-of-thought (CoT) manner \citep{wei2023chainofthoughtpromptingelicitsreasoning}, $E_{sa}$ denotes multiple exemplars. $Rat$ denotes the rationales that represent the thought process and underpin the generation of $P_q$, where both are concurrently produced from a single process.

\subsection{Abstraction-driven Retrieval Module} \label{Section_4_4}
The Abstraction-driven Retrieval module retrieves question-relevant triplets as the evidence to support the answer generation. It helps reduce noise data while decreasing the context length input into the LLM.

Given a question $q$, $n$ initial topic entities $E_q$ and the knowledge graph $KG$, the retrieval phase over the KG is organized into multiple iterative cycles (retrieval iterations). We define two retrieval parameters: width $W$ and depth $D$. Width $W$ denotes the maximum number of topic entities and retrieved relations at each iteration, and depth $D$ represents the maximum number of iterations. 
Each iteration explores the 1-hop neighbor subgraphs of the topic entities and updates them with the discovered neighbors.

After retrieving $n \times W$ relevant entity-relation pairs through relation retrieval, 
previous RAG-based methods use LLMs to retrieve additional entities but perform poorly in the privacy-protected scenario. Instead, we first construct candidate triplets in the Relation-centric Abstraction module. Then, we retrieve question-relevant triplets and identify newly discovered entities within these triplets based on their semantic similarity to the triplets in the aforementioned abstract concept paths.

Specifically, as shown in Part (c) of Figure \ref{fig:fig2}, after the Relation-centric Abstraction module, we gather a set of candidate triplets $T_q^{abs}$. Then, we retrieve $W$ most relevant triplets $T_{q,\text{opt}}^{abs}$ from this set, based on their cosine similarity with the triplets $t_p$ in the abstract concept path $P_q$. The procedure is outlined as follows.

{
\small
\begin{equation}
T_{q,\text{opt}}^{abs} = \underset{\substack{S \subseteq T_q^{abs}, |S| \leq W}}{\arg\max}\; \sum_{t^{abs} \in S} \max_{t_p \in P_q} \cos(\text{Emb}(t^{abs}), \text{Emb}(t_p)))
\label{eq:formu5}
\end{equation}
}

In each retrieval iteration, we retrieve question-relevant triplets $T^{abs}_{q,opt}$ and aggregate all triplets to date into a cumulative set $T^{abs}_{q,all}$. This set is then fed into the Generator module. Additionally, the topic entities are replaced with the newly identified entities from $T^{abs}_{q,opt}$ for the next retrieval iteration.

\subsection{Generator Module}
The Generator module infers the answers by feeding the question and retrieved relevant triplets into the LLM-based generator, which is similar to that of \citet{sun2024thinkongraph}.

Specifically, as shown in Part (d) of Figure \ref{fig:fig2}, after the Abstraction-driven Retrieval module, we gather a set of question-relevant triplets $T^{abs}_{q,all}$ as evidence. Next, we input the evidence, along with the original question $q$, the instruction $Inst_{g}$ and several exemplars $E_{g}$ into an LLM-based generator. The generator then produces a flag $Flag$ and the final answers $Ans$. The process is illustrated as follows.

\begin{equation}
Flag, Ans = \text{LLM}(T^{abs}_{q,all}, q, Inst_{g}, E_{g}).
\label{eq:formu6}
\end{equation}

We output the answers directly if $Flag$ is positive; otherwise, we proceed to the next retrieval iteration. These answers may contain entities anonymized as MIDs. 
These MIDs are replaced with their real names on the user side, which saves a privacy map linking MIDs to their names.

\section{Experiment}

\subsection{Experimental Setup}
\textbf{Datasets} We conduct experiments on three KGQA datasets including WebQSP, CWQ and GrailQA.
The details of the datasets' splits and the number of selected samples are represented in Table \ref{table_datasets}. {WebQSP} \citep{yih-etal-2016-value} contains questions from WebQuestions \citep{berant-etal-2013-semantic} that are answerable by Freebase \citep{Bollacker-etal-2008-freebase}. {CWQ} \citep{talmor-berant-2018-web} extends WebQSP, where most questions require at least 2-hop reasoning. {GrailQA} \citep{Gu-etal-2021-grailqa} is a diverse KGQA dataset, where most questions necessitate long-tail knowledge.

\begin{table}[tb]
  \centering
   \setlength{\tabcolsep}{1.2mm}
{
  \begin{tabular}{cccc}
    \toprule
    {Dataset}  &{\#(Train/Dev/Test)} & {\#Total} & {\#Filtered}\\
    \midrule
    WebQSP & 3,098/-/1,639 & 1,639 & 535\\
    CWQ & 27,639/3,519/3,531 & 1,000  & 449 \\
    GrailQA & 44,337/6,763/13,231 & 1,000 & 645 \\
    \bottomrule
  \end{tabular}
}
  \caption{Statistics of the datasets.}
  \label{table_datasets}
\end{table}

\textbf{Experimental Settings} We define two experimental settings. The first is the \textbf{\#Total setting}, where we conduct the test on our complete sample set. The second is the \textbf{\#Filtered setting}, which simulates a strict privacy-protected scenario. In this scenario, answering questions requires access to private knowledge not included in the LLM’s pre-trained data. To model this scenario, we remove all samples that can be answered correctly via CoT prompting and conduct the test on the remaining filtered subset.

\textbf{Metrics} We use exact match accuracy (Hits@1) as evaluation metric, consistent with previous studies \citep{li-etal-2023-shot,sun2024thinkongraph,NEURIPS_2024_POG}. 

\textbf{Implementation Details} We employ \textsc{gpt-4o-mini-2024-07-18} as the underlying LLM. For the Structure-oriented Abstraction module and the Generator model, the temperature parameter is set to 0 to ensure deterministic outputs. For the Relation-centric Abstraction module, the temperature is set to 0.4 to introduce variability. Both the frequency penalty and presence penalty are set to 0. The width $W$ and depth $D$ of the retrieval are set to 3, striking a balance between performance and efficiency. We repeat the experiment 3 times and report the average scores. 

\subsection{Baselines}

Methods for applying LLMs to the KGQA task are generally divided into three categories by existing studies: Pure-LLM, SP-based and RAG-based methods. To ensure a comprehensive evaluation, we select SoTA methods from each category as our baselines, which are listed as follows.

\textbf{Pure-LLM methods} We select IO \citep{NEURIPS-2020-io}, CoT \citep{wei2023chainofthoughtpromptingelicitsreasoning} and CoT-SC \citep{wang2023selfconsistency} as reference. 
Because they operate without external KGs and do not tailor for KGQA, they serve as baselines for evaluating the intrinsic reasoning capabilities of LLMs.

\textbf{SP-based methods} We select KB-BINDER \citep{li-etal-2023-shot} and TrustUQA \citep{Zhang_Jin_Zhu_Chen_Huang_Wang_Hua_Liang_Chen_2025} as representative SP-based methods. For their variants with dynamically exemplars, we claim them as KB-BINDER-R and TrustUQA-R, respectively.

\textbf{RAG-based methods} We select ToG \citep{sun2024thinkongraph}, PoG \citep{NEURIPS_2024_POG} and GoG \citep{xu-etal-2024-generate} as representative RAG-based methods. ToG \citep{sun2024thinkongraph} uses LLMs to retrieve knowledge triplets via a beam search approach. PoG \citep{NEURIPS_2024_POG} enhances the retrieval process through reflection, memory, and adaptive breadth. In a different vein, GoG \citep{xu-etal-2024-generate} leverages LLMs as a flexible knowledge source, complementing the KG.

\subsection{Performance Comparison}

\begin{table}
  \centering
   \setlength{\tabcolsep}{0.7mm}{
{
  \begin{tabular}{lccccccc}
    \toprule

    \multirow{2}{*}{{Type}} &\multirow{2}{*}{{Method}} &\multicolumn{2}{c}{{WebQSP} }&\multicolumn{2}{c}{{CWQ}} &\multicolumn{2}{c}{{GrailQA}} \\

    \cmidrule(lr){3-4}\cmidrule(lr){5-6}\cmidrule(lr){7-8} & & {\#Tot} &{\#Fil}  & {\#Tot} &{\#Fil} & {\#Tot} &{\#Fil}\\
    \midrule
        \multirow{3}{*}{Pure} 
            & IO & \underline{68.6} & 13.8 & 50.9 & 10.9 & 31.5 & 5.8  \\ 
            & CoT & 67.2 & 0.0 & \underline{55.1} & 0.0 & 35.5 & 0.0 \\
            & CoT-SC & 68.3 & 8.9 & 54.2 & 2.0 & 35.0 & 3.3  \\

    \midrule
        \multirow{4}{*}{SP} 
        & KB-BINDER & 15.8 & 16.0 & - & - & 45.0 & 45.3\\ 
        & KB-BINDER-R & 37.0 & 11.7 & - & - & \underline{62.3} & \underline{62.8} \\ 
        & TrustUQA & 47.4 & 48.4 & - & - & - & -  \\
        & TrustUQA-R & 60.9 & \underline{53.1} & - & - & - & -  \\
    \midrule
        \multirow{5}{*}{RAG} 
           &  ToG & 64.9 & 8.2 & 54.1 & 4.9 & 38.9 & 17.0 \\
           &  GoG & 62.3 & 26.9 & 48.1 & \underline{16.5} & 29.1 & 12.4 \\
           &  PoG & 61.4 & 12.6 & 49.8 & {16.3} & 49.7 & 36.1 \\
            \cmidrule(lr){2-8}
           &  ARoG & \textbf{74.7} & \textbf{58.9} & \textbf{60.0} & \textbf{36.3} & \textbf{78.7} & \textbf{71.8}  \\          
           &  Imp. & \textbf{$\uparrow$6.1} & \textbf{$\uparrow$5.8}& \textbf{$\uparrow$4.9}& \textbf{$\uparrow$19.8}& \textbf{$\uparrow$16.4}& \textbf{$\uparrow$9.0} \\
    \bottomrule
  \end{tabular}  
}  
\caption{Comparison results (\%) of Pure-LLM methods (Pure), SP-based methods (SP), and RAG-based methods (RAG), under the \#Total setting (\#Tot) and \#Filtered setting (\#Fil). ``-'': the SP-based method cannot be run due to lacking special formal query annotations. ``Imp.'': the absolute improvement of ARoG relative to the second-best performer. \textbf{Bold}: the best. \underline{Underline}: the second best. }
    \label{table_2}
}

\end{table}

We compare ARoG with the baselines to demonstrate the effectiveness for question answering, the results are represented in Table \ref{table_2}. Overall, ARoG achieves the best performance across all three datasets under both settings.

First, ARoG significantly outperforms all RAG-based baselines. Existing RAG methods struggle with KGs containing anonymous entities, as they rely heavily on specific entity information for effective retrieval and reasoning.

Second, ARoG also surpasses existing SP-based baselines. This is because ARoG leverages LLMs to directly participate in the reasoning process through iterative retrieval-and-generation mechanisms. In contrast, SP-based baselines lack such iterative paradigms and are constrained by the coverage and completeness of the KG.

Third, compared to Pure-LLM baselines, most RAG-based baselines show no advantage, primarily due to their inability to effectively retrieve entity information in the privacy-protected scenario. In contrast, ARoG demonstrates significant improvements by effectively retrieve question-relevant triplets from the KG.

Finally, from the \#Total setting to the \#Filtered setting, the internal knowledge of LLMs is not enough to answer questions. Therefore, the performance of most RAG-based methods drops significantly, since they fail to retrieve private knowledge from the KG. However, ARoG compensates for this limitation with two abstraction strategies, which helps the LLMs perform reasoning over KGs without concrete entity information. For SP-based methods, since they do not require concrete entity information from the KG, their performance remains relatively stable as well. 


\subsection{Ablation Study}

To assess the effectiveness of the two abstraction strategies in ARoG, we perform ablation studies on three datasets. Specifically, ablation studies are conducted separately on the three steps of the Relation-centric Abstraction module and on the Structure-oriented Abstraction module. Notably, relation retrieval (Step 1) aims to retrieve relations and serves as the foundation for all subsequent steps; relation filtering (Step 2) acts as an enhancement to entity abstraction (Step 3). Therefore, in the ablation studies, we do not consider the scenarios of removing Step 1 or removing Step 3 while retaining Step 2. The results are shown in Table~\ref{table_3}.

\begin{table}[htp]

\setlength{\tabcolsep}{0.9mm}
    {
\begin{tabular}{ccc|c|cc|cc|cc}
\toprule
\multicolumn{4}{c}{ARoG} & \multicolumn{6}{c}{KGQA} \\
\cmidrule(lr){1-4} \cmidrule(lr){5-10}
\multicolumn{3}{c}{RA}   & \multicolumn{1}{c}{SA}    & \multicolumn{2}{c}{WebQSP}                        & \multicolumn{2}{c}{CWQ}                     & \multicolumn{2}{c}{GrailQA}    
\\
\midrule
Step1  & Step2  & Step3  & SA & \#Tot                & \#Fil                & \#Tot               & \#Fil              & \#Tot                 & \#Fil                \\
\midrule
$\checkmark$      & $\times$      & $\times$      & $\times$     &   63.9                   &                53.6      &               36.4      &          20.7          &                  72.6     &   66.9                   \\
$\checkmark$      & $\times$      & $\times$      & $\checkmark$     & 71.6                 & 52.7                 & 54.5                & 30.1               & 75.2                  & 68.0  
\\
$\checkmark$      & $\times$      & $\checkmark$      & $\times$     & 68.3                 &  57.4              & 47.9              & 26.9              & 76.9                  & 70.7        \\
$\checkmark$      & $\times$      & $\checkmark$      & $\checkmark$     & 72.9                 &  \textbf{58.9}                 & 59.5                & 35.4               & 78.3              & 71.6                
\\
$\checkmark$      & $\checkmark$      & $\checkmark$      & $\times$     & 68.1                 & 56.8                 & 50.3                & 30.5              & 76.3                &70.7
\\
$\checkmark$      & $\checkmark$      & $\checkmark$      & $\checkmark$     &  \textbf{74.7}                 &  \textbf{58.9}                 &  \textbf{60.0}                & \textbf{36.3}                & \textbf{78.7}                  &  \textbf{71.8}
\\
\bottomrule
\end{tabular}
}
\caption{Results (\%) of the ablation study. RA: Relation-centric Abstraction module. SA: Structure-oriented Abstraction module. \#Tot: the \#Total setting. \#Fil: \#Filtered setting.}
\label{table_3}
\end{table}

\subsubsection{Analysis on Relation-centric Abstraction Module} \label{chapter5_6}
When omitting the Relation-centric Abstraction module, we observe a performance decline of at least 2.5\% under the \#Total setting and at least 3.8\% under the \#Filtered setting. The decline is most significant under the \#Filtered setting, where answering questions requires acquiring private knowledge not found in the LLM’s pre-trained knowledge. This suggests that the Relation-centric Abstraction module is highly effective at extracting and abstracting in-context semantic information from relations within KG, thereby preserving the private knowledge.

\subsubsection{Analysis on Structure-oriented Abstraction Module}
When omitting the Structure-oriented Abstraction module, we observe a performance decline of at least 2.4\% under the \#Total setting and at least 1.1\% under the \#Filtered setting. The decline is most significant on the CWQ dataset, whose questions primarily require multi-hop reasoning. This suggests that the Structure-oriented Abstraction module excels at analyzing the structural semantics of questions, while the abstract concept paths enhance multi-iteration retrieval.

\subsection{Efficiency Study}
We study the efficiency of RAG-based methods. Table \ref{table_4} reports the average token usage and LLM calls under the \#Total setting. In terms of total token usage, which directly reflects efficiency, ARoG achieves competitive or superior results. On WebQSP, its total token usage is close to ToG and much lower than GoG. On CWQ, ARoG maintains an affordable cost with fewer total token usage than GoG while keeping high performance. 
Notably, on the most challenging GrailQA dataset, ARoG achieves the lowest total token usage, demonstrating clear efficiency advantages in handling complex problems.
Although ARoG requires more LLM calls in some cases, this metric is not directly related to efficiency, as the cost of individual calls varies.

\begin{table}[tb]
  \centering
   \renewcommand\arraystretch{0.9}
    \setlength{\tabcolsep}{0.9mm}
{
  \begin{tabular}{lccccc}
    \toprule
    Dataset&Method	&\#Input	&\#Output	&\#Total  &\#Call \\

    \midrule
    \multirow{4}{*}{WebQSP} 
       & ToG  & 	{6,356.4}	& 1,357.0& 	{7,713.4}& 15.4  \\ 
       & GoG  & 	12,031.0	& {343.3}& 	12,374.3  & {10.6}\\ 
      &  PoG  &	{5,872.5} &	{336.0} &	{6,208.6} &{10.1}\\
       & ARoG & 7,044.7& 	707.3& 	 7,752.1  & 17.3 \\
    
    \midrule
    \multirow{4}{*}{CWQ} 
       & ToG 	&  {8,372.3}	& 1,852.2& 	{10,224.5}  & 21.7\\ 
       & GoG &  14,657.0	& {463.8}& 	15,120.8  & {13.5}	\\ 
      &  PoG 	& {8,730.6}	& {423.2}& 	 {9,153.7}	& {16.9}\\
       & ARoG 	& 10,187.0	& 1,055.0& 	11,242.1  & 25.4\\
        
    \midrule
    
    \multirow{4}{*}{GrailQA} 
       & ToG  & 	{5,220.1}	& 1,306.5& 	6,526.5 & 13.5 \\ 
       & GoG  & 	12,665.7	& 428.0& 	13,093.7 & {11.4} \\ 
      &  PoG  & 	7,445.0	& {436.4} & 	7,881.4  & 15.8 \\
       & ARoG & 	{5,044.7}	& {560.7}& 	{5,605.5}  & {12.5}  \\
    \bottomrule
  \end{tabular} } 
  \caption{Efficiency comparison between RAG-based methods under the \#Total setting. \#Input, \#Output, \#Total represent the number of input tokens, output tokens, and total tokens, respectively. \#Call denotes the number of LLM calls.}
  \label{table_4}
\end{table}

Overall, ARoG demonstrates strong token efficiency across datasets, indicating effective and privacy-protected reasoning with affordable computational cost.



\subsection{Quantitative Study}

\begin{figure}[tb]
  \centering
  \includegraphics[width=\columnwidth]{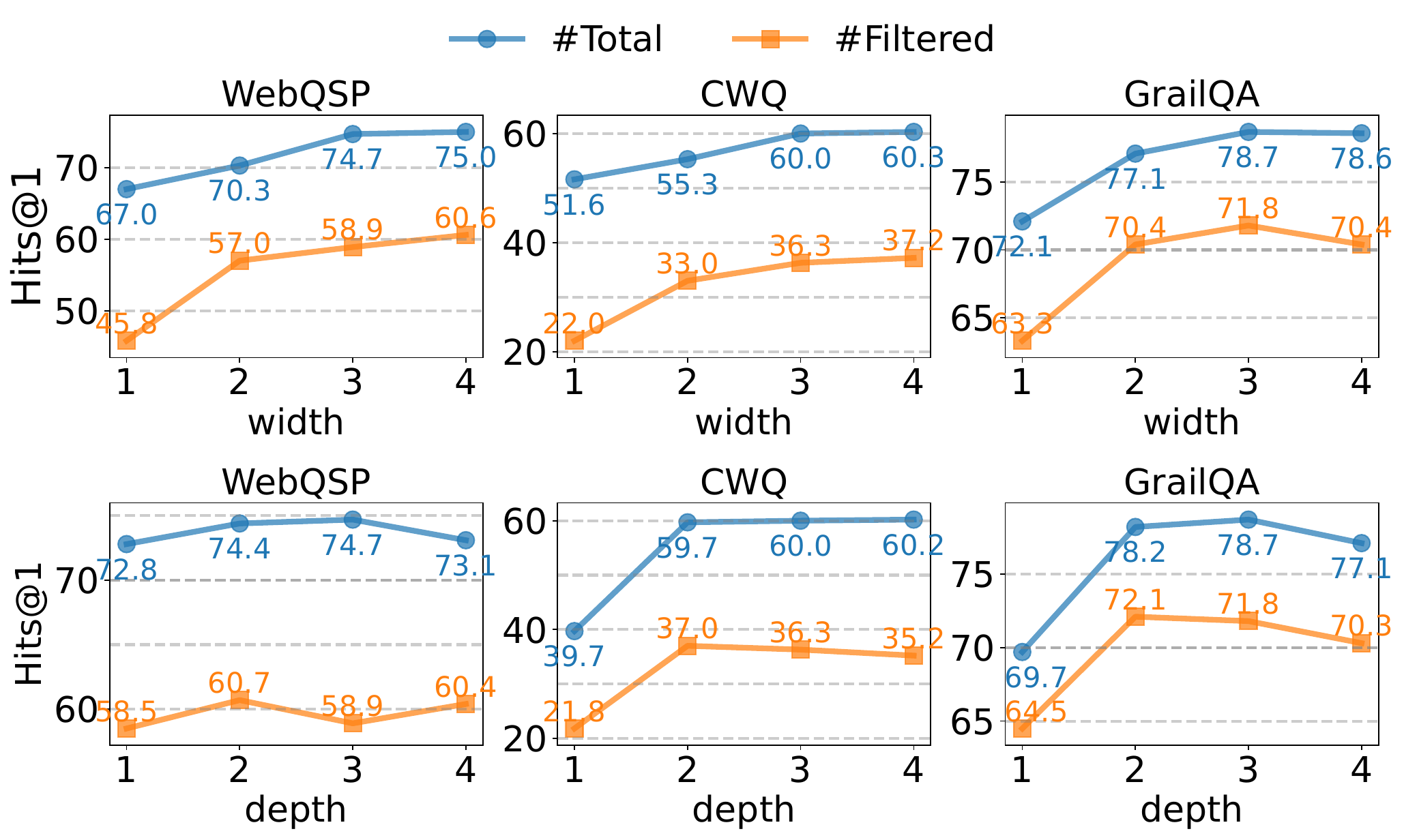}
  \caption{Performance (\%) of ARoG with different retrieval widths and depths.}
  \label{fig:Performance under different retrieval widths and depths}
\end{figure}

To understand the impact of retrieval depth $D$ and width $W$ on ARoG’s performance, we conduct the quantitative study. We vary these parameters from 1 to 4 (with a default of 3) across three datasets, following the approach in ToG \citep{sun2024thinkongraph}. The results are presented in Figure \ref{fig:Performance under different retrieval widths and depths}.

We observe that, on the WebQSP dataset, the width $W$ is particularly critical, while the depth $D$ has little to no impact, as most answer entities can be found within the 1-hop subgraph of initial topic entities. Conversely, for the CWQ and GrailQA datasets, both width and depth enhance performance; however, the benefits diminish beyond a depth of 2, since most answers are located within the 2-hop subgraph.

\subsection{Deep Analysis}
\subsubsection{Comparison between Different Abstraction Strategies}
\begin{figure}[htp]
  \centering
  \includegraphics[width=\columnwidth]{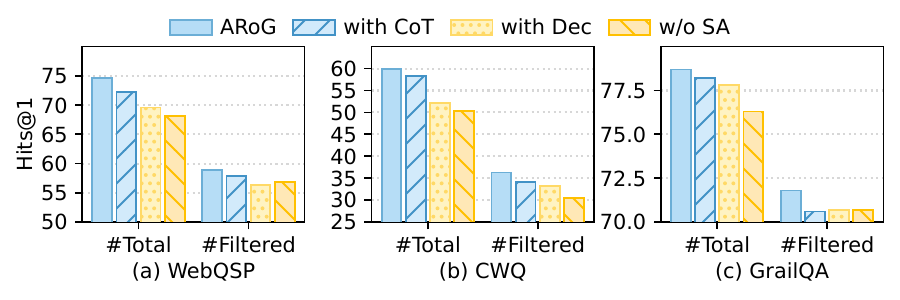}
  \caption{Performance (\%) with different abstraction strategies for the Structure-oriented Abstraction module.}
  \label{fig:Performance with different strategies for Structure-oriented Abstraction}
\end{figure}

To thoroughly validate the effectiveness of abstraction strategy in the Structure-oriented Abstraction module, we compare ARoG against two alternative abstraction strategies as follows. (1) Chain-of-Thought (CoT) \citep{wei2023chainofthoughtpromptingelicitsreasoning}, which replaces the abstract concept path with generated rationales, and (2) question decomposition (Dec) \citep{NEURIPS_2024_POG}, which uses derived sub-questions. For a more comprehensive baseline, we also include the variant that omits the Structure-oriented Abstraction module (``w/o SA''). The performance of all these methods is illustrated in Figure~\ref{fig:Performance with different strategies for Structure-oriented Abstraction}.

We observe that ARoG demonstrates superior performance across all conditions. The CoT succeeds by providing necessary entities, while the Dec succeeds by leveraging structured sub-questions. However, both strategies fail to achieve optimal results since they do not incorporate abstract concepts, which are crucial for the retrieval.

\subsubsection{Performance in Different Privacy-Protected Scenarios}

To further showcase ARoG's robustness, we establish four different privacy scenarios: full privacy-protected scenarios (Private), exposing grounded entities information in {retrieval phase} (P-R), in {generation phase} (P-G), or both phases (P-RAG). We select ToG to represent previous RAG-based methods. We develop ARoG-R that uses the LLMs to retrieve question-relevant entities, thereby replacing the Abstraction-driven Retrieval module in ARoG. The experimental results are illustrated in Figure \ref{fig:Performance of ARoG, ARoG-R and ToG under different privacy-protected phases}.

\begin{figure}[tb]
  \centering
  \includegraphics[width=\columnwidth]{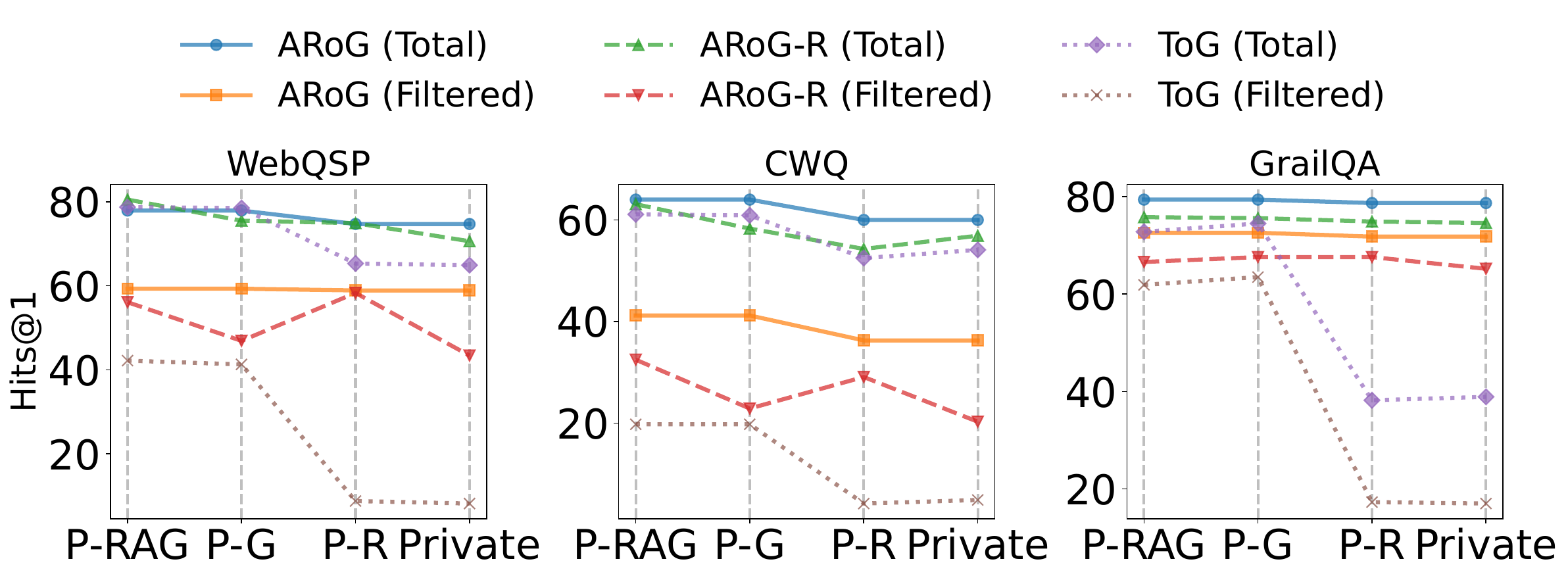}
  \caption{Performance (\%) of ARoG, ARoG-R and ToG in different privacy-protected scenarios.}
  \label{fig:Performance of ARoG, ARoG-R and ToG under different privacy-protected phases}
\end{figure}

We observe that, when entities are anonymous in the {generation phase}, ToG's performance drops significantly while ARoG and ARoG-R remain. The reason is that the anonymous entities retrieved by ToG are meaningless, while ARoG and ARoG-R address this issue based on the relation-centric abstraction. Additionally, when entities are anonymous in the retrieval phase, ARoG-R drops significantly while ARoG remains high. Although ToG also maintains its performance, it shows a relatively poor result. The reason is that ToG and ARoG-R retrieve the anonymous entities by feeding the question along with anonymous entities to LLMs. This approach results in suboptimal performance. In contrast, ARoG retrieve the question-relevant anonymous entities based on the Abstraction-driven Retrieval module.

Overall, ARoG exhibits impressive robustness and high performance across all three datasets and scenarios.


\section{Conclusion}
In this paper, we investigate the \emph{privacy-protected RAG scenario} in KGQA for the first time,
where the entities within KGs are anonymous to LLMs. To address the challenges in this scenario, we propose a novel ARoG framework that incorporates relation-centric and structure-oriented abstraction strategies. Based on these two strategies, ARoG is able to effectively retrieve question-relevant knowledge triplets while preventing data privacy. 
Extensive experiments conducted on three datasets demonstrate the effectiveness and robustness of ARoG, highlighting its potential to address the challenging \emph{privacy-protected RAG scenario} for KGQA.

\section*{Acknowledgments}
This work was supported by the National Natural Science
Foundation of China (NSFC) (Grant No. 62576256), the
Fundamental Research Funds for the Central Universities,
China (Grant No. 2042022dx0001), and the Key Laboratory of Computing Power Network and Information Security, Ministry of Education (Grant No. 2024ZD027).

\bibliography{aaai2026}

\clearpage

\end{document}